\documentclass[runningheads]{llncs}

\usepackage{eccv}

\usepackage{eccvabbrv}

\usepackage{graphicx}
\usepackage{booktabs}
\usepackage{mathtools}
\usepackage{stfloats}
\usepackage{pifont}
\usepackage{dsfont}

\newcommand{\cmark}{\ding{51}}
\newcommand{\xmark}{\ding{55}}

\usepackage{xcolor}
\definecolor{red}{rgb}{1,0.2,0.2}

\usepackage[accsupp]{axessibility}  

\usepackage{hyperref}

\usepackage{orcidlink}
\usepackage{ulem}
\usepackage{bm}
\usepackage{enumitem}
\usepackage{upgreek}

\begin{document}

\title{AA-Splat: Anti-Aliased Feed-forward 3D Gaussian Splatting}

\titlerunning{AA-Splat}

\author{Taewoo Suh\inst{1} \and
Sungpyo Kim\inst{1} \and
Jongmin Park\inst{1}\orcidlink{https://orcid.org/0009-0002-8757-0139} \and
Munchurl Kim\inst{1}\orcidlink{https://orcid.org/0000-0003-0146-5419}}

\authorrunning{T. Suh et al.}

\institute{Korea Advanced Institute of Science and Technology\\
\email{\{twsuh, ksp04204, jm.park, mkimee\}@kaist.ac.kr}
https://kaist-viclab.github.io/aasplat-site}

\maketitle

\begin{center}
\includegraphics[width=1.0\linewidth]{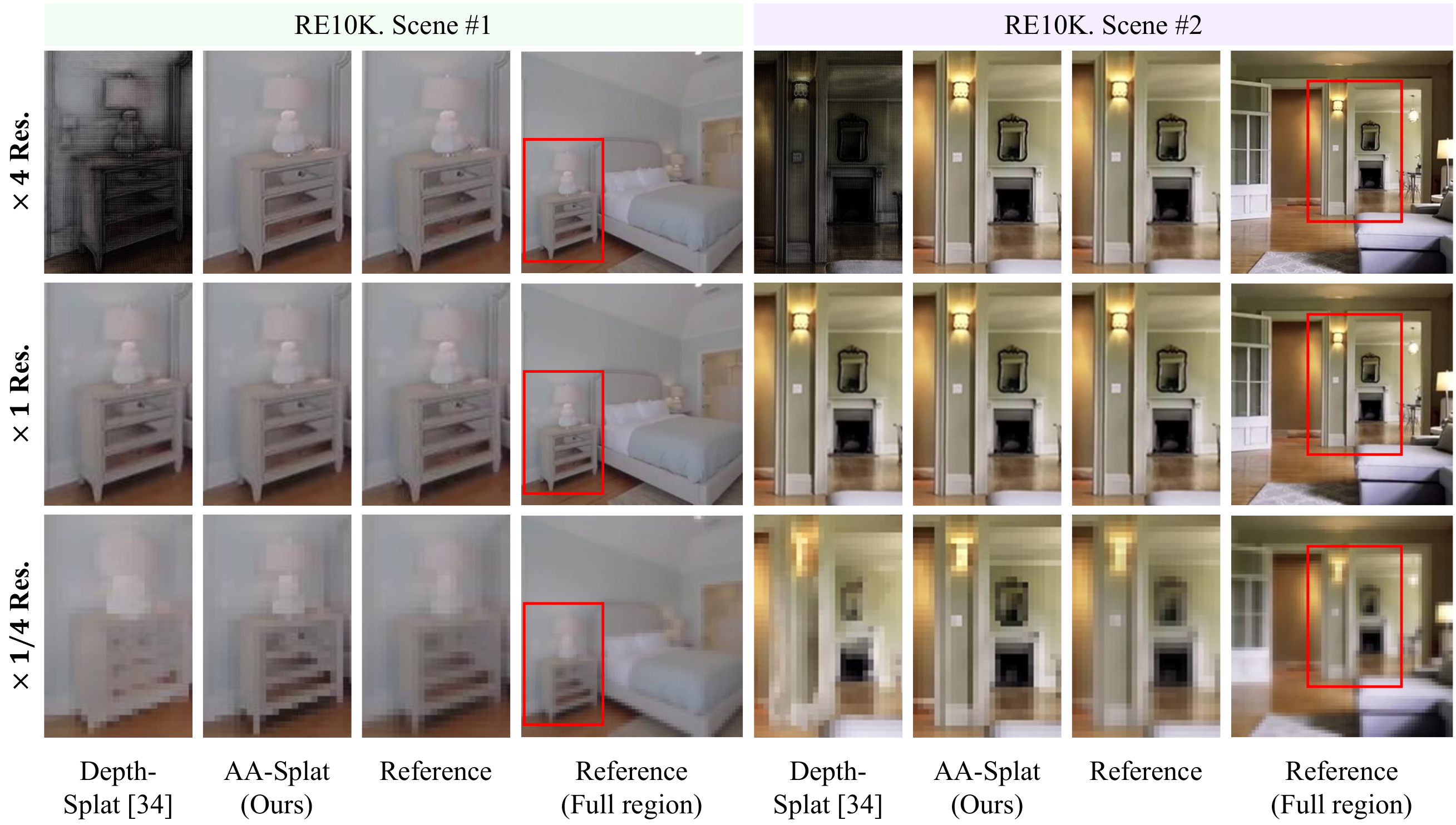}
\captionof{figure}{Our AA-Splat is the first feed-forward 3DGS framework for alias-free rendering. Built upon the DepthSplat \cite{xu2025depthsplat} architecture, AA-Splat employs Opacity Balanced Band Limiting (OBBL) combining 3D band-limiting post-filter (3D-BLPF) and opacity balancing (OB). This enables robust, high-fidelity reconstruction at out-of-distribution scales (1/4× to 4× resolution). While the baseline DepthSplat \cite{xu2025depthsplat} suffers from severe erosion and dilation artifacts when zooming (changing sampling rates), AA-Splat consistently produces sharp, anti-aliased results.}
\label{fig:teaser}
\end{center}

\begin{abstract}
    Feed-forward 3D Gaussian Splatting (FF-3DGS) emerges as a fast and robust solution for sparse-view 3D reconstruction and novel view synthesis (NVS).
    However, existing FF-3DGS methods are built on incorrect screen-space dilation filters, causing severe rendering artifacts when rendering at out-of-distribution sampling rates.
    We \textit{firstly} propose an FF-3DGS model, called AA-Splat, \textit{to enable robust anti-aliased rendering at any resolution}.
    AA-Splat utilizes an opacity-balanced band-limiting (OBBL) design, which combines two components: a 3D band-limiting post-filter integrates multi-view maximal frequency bounds into the feed-forward reconstruction pipeline, effectively band-limiting the resulting 3D scene representations and eliminating degenerate Gaussians; an Opacity Balancing (OB) to seamlessly integrate all pixel-aligned Gaussian primitives into the rendering process, compensating for the increased overlap between expanded Gaussian primitives.
    AA-Splat demonstrates \textit{drastic improvements} with average 5.4$ \sim $7.5dB PSNR gains on NVS performance over a state-of-the-art (SOTA) baseline, DepthSplat, at all resolutions, between $4\times$ and $1/4\times$. Code will be made available.
\end{abstract}

\section{Introduction}
\label{sec:intro}
3D reconstruction from multi-view images and novel view synthesis (NVS) are vital computer vision tasks, with applications including virtual and augmented reality, robotics, and autonomous driving. A milestone in NVS was achieved with the advent of 3D Gaussian Splatting (3DGS) \cite{kerbl20233d}, which represents 3D scene contents as collections of differentiable 3D Gaussian primitives which can be rendered extremely fast by projection to 2D screen space and rasterization, while retaining photorealistic visual fidelity. A more recent emerging trend is the feed-forward 3DGS (FF-3DGS) model \cite{charatan2024pixelsplat, szymanowicz2024splatter, chen2024mvsplat, chen2024mvsplat360, xu2025depthsplat, kang2025ilrm, zhang2024gs, tang2024lgm}. While previous `per-scene' 3DGS models \cite{guedon2024sugar, yang2024deformable, wu20244d, lu2024scaffold, fan2024lightgaussian, keetha2024splatam, yan2024gs} optimized single models to represent only one scene, FF-3DGS models use one large-sized network trained on a collection of multiple scenes to directly infer the 3DGS representations of unseen scenes, eliminating the costly optimization process.

However, existing FF-3DGS models \cite{chen2024mvsplat, xu2025depthsplat, ye2024no, huang2025no, huang2025spfsplatv2} suffer from severe unnatural rendering artifacts when rendering sampling rates (e.g., camera viewing distances, focal lengths, resolutions) are greatly altered from those present in context views. This is a form of aliasing, similar in nature to that which had been observed in per-scene 3DGS. 3DGS is fundamentally an inverse problem that aims to recover a continuous representation of a 3D scene. However, the vanilla 3DGS rendering engine \cite{kerbl20233d}, which uses a physically incorrect screenspace dilation filter, causes the recovery of highly degenerate (thin, needle-like) Gaussians. Although these flaws are masked by the dilation filter when rendering at sampling rates similar to those visible during reconstruction, they are revealed as: (i) erosion artifacts where structures appear unnaturally thin when sampling rate is increased (e.g., under zoom-in), and (ii) dilation artifacts where structures appear unnaturally thick when sampling rate is decreased (e.g., under zoom-out). While these flaws have been addressed in per-scene 3DGS methods \cite{yu2024mip, younes2025anti, chen2025alias, steiner2025aaa}, FF-3DGS models continue to be built with the physically inaccurate dilation filters of the vanilla 3DGS \cite{kerbl20233d}. As a result, these models regress highly degenerate (thin) Gaussians, resulting in (i) severe erosion artifacts that open up transparent seams when increasing the sampling rate, and (ii) dilation artifacts of structures when reducing the sampling rate. Adopting a more physically accurate 2D screen-space Mip filter \cite{yu2024mip} resolves such dilation artifacts while closing up most of the false transparency artifacts, but the FF-3DGS models with the 2D Mip filter \cite{yu2024mip} still tend to regress degenerate (overly thin) Gaussians. This necessitates fundamentally band-limiting the 3D Gaussian representation.

To overcome these challenges, we propose AA-Splat, the first Anti-Aliased FF-3DGS model that enables alias-free rendering at varying sampling rates. First, we propose to integrate the multi-view frequency bounds \cite{yu2024mip} based on the Nyquist sampling rate into the FF-3DGS architecture in the 3D Band-limiting Post-Filtering (3D-BLPF). However, we observe that this alone results in subpar NVS performance as well as unnatural grid-like artifacts when zooming in, due to the application of a 3D filter expanding pixel-aligned Gaussian primitives and increasing the overlap between them, thus decreasing the visibility of more distant Gaussians. We thus combine 3D-BLPF with an opacity balancing (OB) compensation technique to seamlessly integrate all Gaussian primitives into the rendering process and reduce unnatural grid-like artifacts. The resulting combination, opacity-balanced band-limiting (OBBL), effectively produces renegerate 3D Gaussian representations and enables alias-free rendering across a wide range of sampling rates (e.g., zoom levels). Fig. \ref{fig:teaser} demonstrates the anti-aliasing effects by our AA-Splat. The baseline model, DepthSplat \cite{xu2025depthsplat}, suffers from severe erosion artifacts which manifest as black voids and seams between pixel-aligned primitives when rendering at higher resolution, and significant dilation artifacts with various structures rendered as being unnaturally thick when rendering at lower resolutions. On the other hand, AA-Splat effectively suppresses aliasing artifacts under drastic changes to sampling rates, producing more naturally down-sampled or smoothly interpolated novel views. To summarize, our contributions are as follows:
\begin{itemize}
    \item We introduce AA-Splat, the first feed-forward 3D Gaussian Splatting model that enables alias-free rendering under drastic changes to sampling rates such as zooming in or zooming out;
    \item We design the opacity-balanced band-limiting (OBBL) to reconstruct regenerate Gaussians using a combination of two complementary techniques,
    a 3D band-limiting post-filter (3D-BLPF) that effectively band-limits the 3D Gaussian primitives and eliminates degenerate Gaussians,
    and opacity balancing (OB) that compensates for the resulting increase in overlap between pixel-aligned Gaussian primitives and seamlessly integrates all Gaussians into the rendering process;
    \item We demonstrate \textit{significantly improved} NVS performance with average 5.4 $\sim$ 7.5dB PSNR gains when applied on existing SOTA feedforward methods under drastic rendering resolution variations ($1/4\times$ $\sim$ $4\times$) on common NVS benchmarks.
\end{itemize}

\section{Related Works}
\label{sec:rel-works}
\subsection{Per-Scene Optimization-based Novel View Synthesis}
\label{sec:rel-works-per-scene-nvs}
The advent of Neural Radiance Fields (NeRFs) \cite{mildenhall2021nerf} was a major turning point for both 3D scene reconstruction and novel view synthesis (NVS) research, thanks to their photorealistic high-fidelity rendering capabilities.
This potential has since inspired a wide variety of follow-up works \cite{bui2025moblurf,cao2023hexplane,chen2022tensorf,fridovich2022plenoxels,fridovich2023k,garbin2021fastnerf,hu2022efficientnerf,muller2022instant,yu2021plenoctrees}.
More recently, 3D Gaussian Splatting \cite{kerbl20233d} proposed to represent a scene as a collection of 3D Gaussian primitives, which can efficiently and quickly render photorealistic novel views by rasterization-based differentiable rendering. 3DGS can also reconstruct the scene from a given set of training views by photometric supervision and optimization, as the entire rendering pipeline is fully differentiable. This efficiency and high visual fidelity has made 3DGS the centerpiece of numerous follow-up works \cite{guedon2024sugar,yang2024deformable,wu20244d,lu2024scaffold,fan2024lightgaussian,keetha2024splatam,yan2024gs}.
However, 3DGS relies on per-scene optimization for reconstruction, and cannot generalize to unseen scenes without retraining. This results in slow reconstruction, as well as a limited ability to faithfully reconstruct scenes from sparse input/training views due to overfitting.

\subsection{Feed-forward 3D Gaussian Splatting}
\label{sec:rel-works-ff-3dgs}
A more recently emergent trend is the feed-forward 3DGS (FF-3DGS) model \cite{charatan2024pixelsplat, szymanowicz2024splatter, chen2024mvsplat, chen2024mvsplat360, xu2025depthsplat, kang2025ilrm, zhang2024gs, tang2024lgm}. Unlike previous `per-scene' 3DGS which optimized a single model to represent only one scene, the FF-3DGS models use a large model trained on a collection of multiple scenes to directly infer the 3DGS representations of unseen scenes, eliminating the costly optimization process in favor of a single forward pass of a model. Furthermore, they can also transfer rich data-driven scene priors to unseen scenes, enabling robust reconstruction from sparse views.
MVSplat \cite{chen2024mvsplat} used a compact plane sweeping-based multi-view stereo model to estimate per-view depths that can be unprojected to directly obtain the Gaussian centers. DepthSplat \cite{xu2025depthsplat} extended MVSplat by incorporating rich priors from monocular depth foundation models \cite{yang2024depth}, greatly improving robustness in occluded regions or featureless surfaces where feature matching becomes unreliable. NoPoSplat \cite{ye2024no} used a pose-free geometric transformer \cite{wang2024dust3r, leroy2024grounding} to enable feedforward reconstruction without the need for pre-computed pose inputs, although it still needed ground truth pose labels during training. SPFSplat \cite{huang2025no} and SPFSplatV2 \cite{huang2025spfsplatv2} further expanded on pose-free FF-3DGS by removing the need for pose annotations during training.

\subsection{Anti-Aliased Novel View Synthesis}
\label{sec:rel-works-aa-nvs}
Anti-aliasing in novel view synthesis refers to enabling novel views to be rendered at varying image sampling rates without introducing unnatural artifacts. Rendering a scene from higher resolution, moving the camera closer in, or increasing the focal length increases the sampling rate. On the other hand, rendering from lower resolution, moving the camera further out, or decreasing the focal length decreases the sampling rate.
In NeRFs, anti-aliasing was implemented via cone tracing and pre-filtering of MLP input positional encodings so that both components represented volumes in space rather than the infinitesimal lines or points\cite{barron2021mip,barron2022mip}.
In per-scene 3DGS, Mip-Splatting \cite{yu2024mip} identified the erosion and dilation artifacts in vanilla 3DGS \cite{kerbl20233d} and proposed a 3D smoothing filter and a 2D screen-space Mip filter to mitigate them.
AA-2DGS \cite{younes2025anti} modified these filters to be compatible with 2DGS \cite{huang20242d}, which uses planar 2D Gaussian primitives.
AAA-Gaussians \cite{steiner2025aaa} applied Mip-Splatting's anti-aliasing filters to frameworks where Gaussians were evaluated and rendered directly in 3D space without projection to screen-space, while also expanding upon the problem formulation to handle other view-dependent artifacts such as popping or distortion artifacts.
4DAA \cite{chen2025alias} extended Mip-Splatting \cite{yu2024mip} to dynamic scenes, and proposed several modifications to the 3D smoothing filter designs to better reconcile them with the time-dependent nature of dynamic primitives.
However, these methods only target per-scene optimization scenarios. To the best of our knowledge, we are the first to address anti-aliasing in FF-3DGS models.


\section{Preliminaries}
\subsection{3D Gaussian Splatting (3DGS)}
3DGS \cite{kerbl20233d} represents a 3D scene as a collection of 3D Gaussian primitives. The $k$-th Gaussian primitive is defined in 3D world coordinates by its center position (mean vector) $\bm{\upmu}_k\in\mathbb{R}^{3\times1}$, covariance matrix $\mathbf{\Sigma}_k\in\mathbb{R}^{3\times3}$, opacity $\alpha_k\in[0, 1]$, and spherical harmonics coefficients $\mathbf{sh}_k$ for view-dependent color, where the 3D Gaussian function is denoted as:
\begin{equation}
  \begin{split}
  \mathcal{G}_k(\mathbf{x})=\exp\bigg(-\frac{1}{2}(\mathbf{x}-\bm{\upmu}_k)^T\mathbf{\Sigma}_k^{-1}(\mathbf{x}-\bm{\upmu}_k)\bigg)
  \end{split}
  \label{eq:gs_primitive}
\end{equation}
where $\mathbf{\Sigma}_k$ is parameterized as a semi-definite matrix product that is given by $\mathbf{\Sigma}_k=\mathbf{R}_k\mathbf{s}_k\mathbf{s}_k^T\mathbf{R}_k^T$ to ensure that it always remains within the space of valid covariance matrices. Here, $\mathbf{s}_k\in\mathbb{R}^{3\times1}$ is a 3D scaling vector and $\mathbf{R}_k\in\mathbb{R}^{3\times3}$ is a rotation matrix parameterized by a quaternion $\mathbf{q}_k\in\mathbb{R}^4$ \cite{kerbl20233d}.\\
When rendering a 2D image from these 3D Gaussian primitives $\{\mathcal{G}_k\}$ from a view point defined by camera rotation matrix $\mathbf{R}\in\mathbb{R}^{3\times3}$ and translation vector $\mathbf{t}\in\mathbb{R}^{3}$, the 3D Gaussians are first transformed from 3D world coordinates to 3D camera coordinates as:
\begin{equation}
  \begin{split}
  \bm{\upmu}_k^\text{cam}=\mathbf{R}\bm{\upmu}_k+\mathbf{t}, \mathbf{\Sigma}_k^\text{cam}=\mathbf{R}\mathbf{\Sigma}_k\mathbf{R}^T
  \end{split}
  \label{eq:gs_world_to_camera}
\end{equation}
These transformed 3D Gaussians are then projected from 3D camera space to 2D image space via a local affine transformation as:
\begin{equation}
  \begin{split}
  \mathbf{\Sigma}_k^\text{img}=\mathbf{J}_k\mathbf{\Sigma}_k^\text{cam}\mathbf{J}_k^T
  \end{split}
  \label{eq:gs_camera_to_image}
\end{equation}
where $\mathbf{J}_k$ denotes the Jacobian of the affine approximation to the projective transformation. Omitting the third row and column of $\mathbf{\Sigma}_k^\text{img}$ yields the 2D covariance matrix $\mathbf{\Sigma}_k^\text{2D}$ in image space, and we use $\mathcal{G}_k^\text{2D}$ to refer to the corresponding projected 2D Gaussian \cite{kerbl20233d}. After projection, 3DGS \cite{kerbl20233d} models view-dependent color $\mathbf{c}_k$ using spherical harmonics, and renders final image color at $\mathbf{x}$ via alpha blending according to the orders of the primitives' depths as:
\begin{equation}
  \begin{split}  C(\mathbf{x})=\sum_{i\in\mathcal{N}}\mathbf{c}_i\alpha_i\mathcal{G}_i^\text{2D}(\mathbf{x})\prod_{j=1}^{i-1}(1-\alpha_j\mathcal{G}_j^\text{2D}(\mathbf{x}))
  \end{split}
  \label{eq:gs_alpha_blending}
\end{equation}
where $\mathcal{N}$ refers to the Gaussians in proximity of the pixel at position $\mathbf{x}$ to be rendered.

\begin{figure}[tb]
  \centering
  \includegraphics[width=1.0\linewidth]{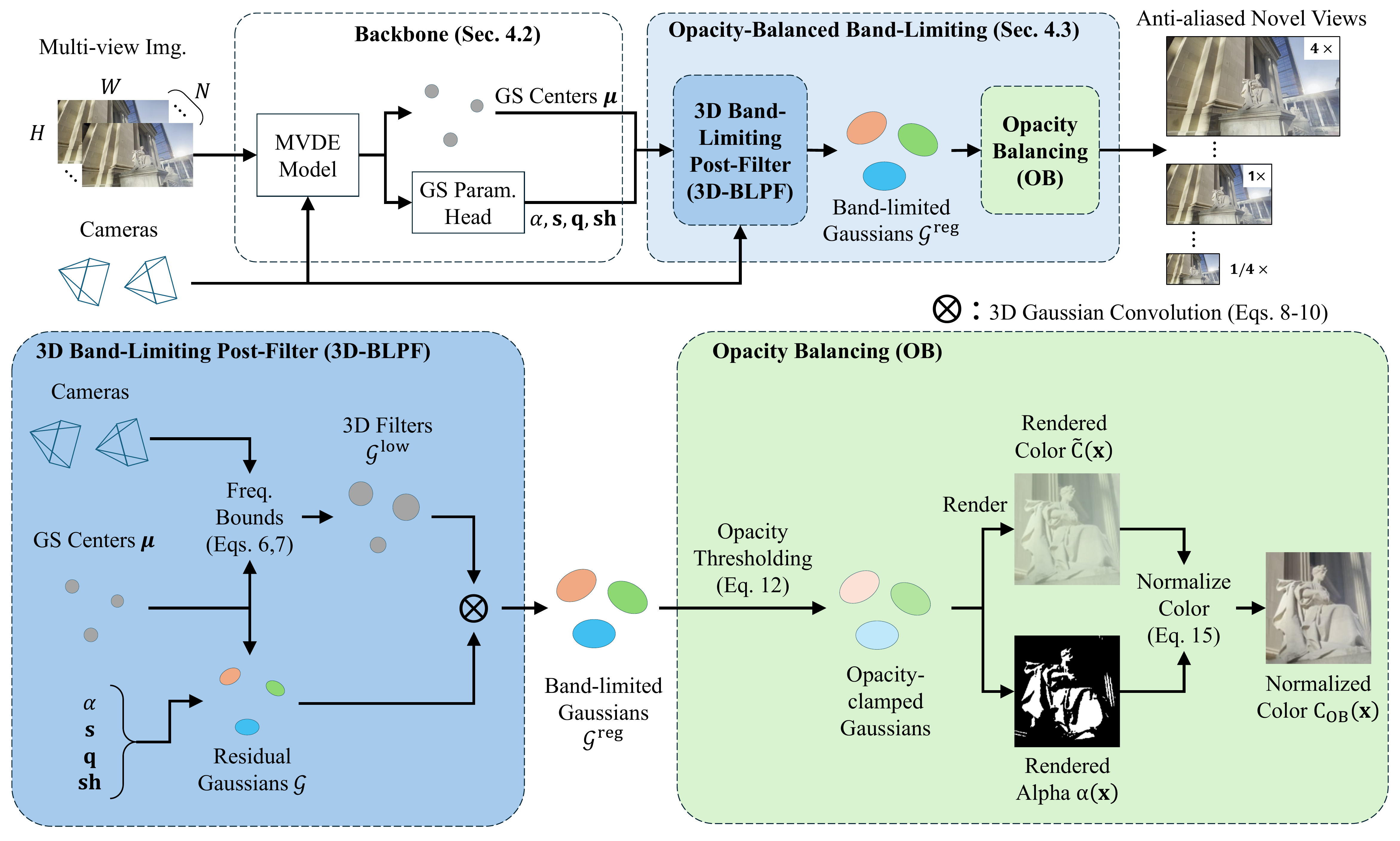}
  \caption{\textbf{Overview of AA-Splat.} After unprojecting per-view depth maps to obtain the 3D Gaussian centers, the maximal frequency bound of each 3D Gaussian primitive is computed from context view camera poses and focal lengths, which is then converted to its 3D band-limiting Gaussian post-filter. The GS head predicts the residual Gaussian parameters which are then post-filtered by the 3D filters to obtain the band-limited 3D Gaussians. During rendering, opacity balancing (OB) compensates for the increase in overlap between pixel-aligned Gaussian primitives for improved photometric supervision and rendering fidelity.}
  \label{fig:main-architecture}
\end{figure}

\section{Proposed Method: AA-Splat}
\subsection{Overview}
Fig. \ref{fig:main-architecture} illustrates the overall framework of our AA-Splat. Given $N$ context view images $\{\mathbf{I}^i\}_{i=1}^N$ ($\mathbf{I}^i\in\mathbb{R}^{H\times W\times 3}$, where $H$ and $W$ are image sizes) and their corresponding camera projection matrices $\{\mathbf{P}^i\}_{i=1}^N$ ($\mathbf{P}^i\in\mathbb{R}^{3\times4}$), AA-Splat aims to predict a set of pixel-aligned Gaussian primitives $\{\mathcal{G}_j\}_{j=1}^{H\cdot W\cdot N}$ that maximizes the rendering quality under drastic changes to sampling rates. That is, we aim to suppress rendering artifacts caused when zooming-in/out, changing the camera viewing distance, or changing the camera focal length. Our proposed AA-Splat computes the minimum scale of each Gaussian primitive $\mathcal{G}_j$ as the Nyquist sampling interval, effectively band-limiting the reconstructed 3DGS representation. As this bandlimiting inherently expands the scales of the 3D primitives, it also increases the overlap caused by pixel-aligned Gaussian primtives, hindering NVS performance. To compensate for this, we adopt an opacity blending (OB) technique to ensure that all Gaussian primitives are seamlessly integrated into the rendering process. This combination not only effectively removes unnatural aliasing artifacts, but also serves as a robust scale-invariant prior that helps improve cross-domain generalization.

In the following sections, we detail our backbone architecture in Sec. \ref{sec:backbone}, followed by our filtering and compensation mechanism in Sec. \ref{sec:combined-filter}.

\subsection{Architecture}
\label{sec:backbone}
As illustrated in Fig. \ref{fig:main-architecture}, the backbone, DepthSplat \cite{xu2025depthsplat}, receives as input $N$ context view images $\{\mathbf{I}^i\}_{i=1}^N$ ($\mathbf{I}^i\in\mathbb{R}^{H\times W\times 3}$, where $H$ and $W$ are image sizes) and their corresponding camera projection matrices $\{\mathbf{P}^i\}_{i=1}^N$ ($\mathbf{P}^i\in\mathbb{R}^{3\times4}$) which are computed from the intrinsic and extrinsic matrices.
DepthSplat \cite{xu2025depthsplat} starts with a robust multi-view depth (DE) estimation model which combines explicit plane-sweeping-based multi-view stereo features \cite{xu2023unifying, chen2024mvsplat} with the rich priors of monocular foundation models \cite{yang2024depth} via channel-wise feature concatenation. This multi-view DE model yields per-view dense depth maps $\{\mathbf{D}^i\}_{i=1}^N$ ($\mathbf{D}^i\in\mathbb{R}^{H\times W}$) and various intermediate features which can be denoted as per-view image features $\{\mathbf{f}^i\}_{i=1}^N$ for simplicity.
This predicted per-view depth map is unprojected into world space to obtain the pixel-aligned Gaussian centers $\{\bm{\upmu}_j\}_{j=1}^{H\cdot W\cdot N}$. A lightweight Gaussian predictor network utilizes $\{\mathbf{f}^i\}_{i=1}^N$ and $\{\mathbf{D}^i\}_{i=1}^N$ to predict the remaining Gaussian parameters: opacities $\{\alpha_j\}_{j=1}^{H\cdot W\cdot N}$, Gaussian scales $\{\mathbf{s}_j\}_{j=1}^{H\cdot W\cdot N}$ and rotation quaternions $\{\mathbf{q}_j\}_{j=1}^{H\cdot W\cdot N}$ which are then used to compute the Gaussian covariance matrices $\{\mathbf{\Sigma}_j\}_{j=1}^{H\cdot W\cdot N}$, and spherical harmonics coefficients $\{\mathbf{sh}_j\}_{j=1}^{H\cdot W\cdot N}$ for computing the view-dependent Gaussian colors $\{\mathbf{c}_j\}_{j=1}^{H\cdot W\cdot N}$. The Gaussian parameters $\{\alpha_j,\mathbf{s}_j,\mathbf{q}_j,\mathbf{sh}_j\}_{j=1}^{H\cdot W\cdot N}$ are obtained at the GS Head:
\begin{equation}
  \begin{split}
  \{\alpha_j,\mathbf{s}_j,\mathbf{q}_j,\mathbf{sh}_j\}_{j=1}^{H\cdot W\cdot N}
  = \text{Head}_\text{GS}(\{\mathbf{D}^i\}_{i=1}^N,\{\mathbf{f}^i\}_{i=1}^N), \\
  \{\mathcal{G}_j\}_{j=1}^{H\cdot W\cdot N}\triangleq\{\bm{\upmu}_j,\alpha_j,\mathbf{s}_j,\mathbf{q}_j,\mathbf{sh}_j\}_{j=1}^{H\cdot W\cdot N}.
  \end{split}
  \label{eq:gs_head}
\end{equation}

A single inference of the feedforward model thus reconstructs the 3D scene as a collection of pixel-aligned Gaussian primitives $\{\mathcal{G}_j\}_{j=1}^{H\cdot W\cdot N}$.
Please note that the backbone architecture of AA-Splat is not restricted to a specific architecture such as the DepthSplat \cite{xu2025depthsplat} but can be applicable to any FF-3DGS model. 

\subsection{Opacity-Balanced Band-Limiting for Regenerate Gaussians}
\label{sec:combined-filter}
Although DepthSplat \cite{xu2025depthsplat} is capable of faithfully reconstructing 3D geometry from a collection of posed context views, it still yields many degenerate Gaussians which are much smaller than single image-space pixels, causing severe erosion and false transparency artifacts when increasing the rendering sampling rate. To remedy this, 3D low-pass post-filtering can be applied for the degenerate Gaussians, as done in `per-scene' anti-aliasing GS of Mip-Splatting  \cite{yu2024mip}.
However, unlike per-scene 3DGS, we observed that naively applying 3D filtering to FF-3DGS models causes unnatural grid-like artifacts when sampling rates are increased (Fig. \ref{fig:abl-dl3dv-qualitative}). This is due to the band-limiting operation that forcibly expands the Gaussians, thus causing (i) the excessive overlap between pixel-aligned 3D Gaussian primitives and (ii) the decreased visibilities for farther Gaussians that tend to be excluded (ignored) from photometric supervision.
To mitigate this, in addition to applying the 3D low-pass post-filtering for the degenerate Gaussian, we propose an Opacity-Balanced Band-Limiting (OBBL) technique to obtain regenerate Gaussians. We first briefly describe the 3D band-limiting post-filter \cite{yu2024mip}, and explain an opacity-balancing compensation technique based on forced alpha blending \cite{he2026surfsplat}, both of which, in combination, help yield regenerate Gaussians for alias-free rendering.

\subsubsection{3D Band-Limiting Post-Filtering (3D-BLPF).}

Starting from the predicted world-space Gaussian centers $\{\bm{\upmu}_j\}_{j=1}^{H\cdot W\cdot N}$, we project each Gaussian to the $N$ context views. We denote the screen-space visibility of $\mathcal{G}_j$ at view $i$ as $\mathds{1}^{(i)}(\mathbf{\mu}_j)$, its depth at view $i$ as $d_j^{(i)}$, and the camera focal length of view $i$ as $f^{(i)}$. The maximal sampling rate for primitive $\mathcal{G}_j$ is determined as \cite{yu2024mip}:
\begin{equation}
    \hat{\nu}_j = \max\big(\{\mathds{1}^{(i)}(\bm{\upmu}_j) \cdot f^{(i)} / d_j^{(i)}\}_{i=1}^N\big).
    \label{eq:maximal-sampling-rate}
\end{equation}
We then compute the 3D Band-Limiting Post-Filter (3D-BLPF) $\mathcal{G}_j^\text{low}$ corresponding to the Gaussian primitive $\mathcal{G}_j$ as a spherical isotropic 3D Gaussian, which will serve as the minimum scale of its band-limited counterpart $\mathcal{G}_j^\text{reg}$:
\begin{equation}
    \mathcal{G}_{j}^\text{low}(\mathbf{x})=\exp{\bigg(-\frac{1}{2}(\mathbf{x}-\bm{\upmu}_j)^T\Big(\frac{\sigma_s}{\hat{\nu}_j}\mathbf{I}\Big)(\mathbf{x}-\bm{\upmu}_j)\bigg)},
    \label{eq:gaussian-lpf}
\end{equation}
where $\sigma_s$ is a filter size hyperparameter which is empirically set to $0.2$.

We obtain the band-limited Gaussian primitives by repurposing the GS Head to predict the residuals of the Gaussian parameters, such that the 3D-BLPF can be applied to the raw predicted Gaussians by 3D convolution:
\begin{equation}
    \mathcal{G}_{j}^\text{reg}(\mathbf{x})=(\mathcal{G}_{j}\otimes\mathcal{G}_{j}^\text{low})(\mathbf{x}).
    \label{eq:3d-conv}
\end{equation}
where $\otimes$ is a 3-D convolution operator.
The scale vector of the final band-limited (smoothed) Gaussian primitives can be expressed as:
\begin{equation}
    \mathbf{s}_{j}^\text{reg}=[s_{j,x}^\text{reg},s_{j,y}^\text{reg},s_{j,z}^\text{reg}]^T,
    \label{eq:filtering-simplified}
\end{equation} 
where, by letting  $s'_j=\sigma_s/\hat{\nu_j}$, we have
\begin{equation}
    s_{j,x}^\text{reg}=\sqrt{s_{j,x}^2+{s'_j}^2}, \\
    s_{j,y}^\text{reg}=\sqrt{s_{j,y}^2+{s'_j}^2},
    s_{j,z}^\text{reg}=\sqrt{s_{j,z}^2+{s'_j}^2}.
    \label{eq:filtering-simplified-2}
\end{equation} \\

The parameters of our final band-limited Gaussian primitives $\{\mathcal{G}_j^\text{reg}\}_{j=1}^{H\cdot W\cdot N}$ can be expressed as:
\begin{equation}
    \{\mathcal{G}_j^\text{reg}\}_{j=1}^{H\cdot W\cdot N}\triangleq\{\bm{\upmu}_j,\alpha_j,\mathbf{s}_j^\text{reg},\mathbf{q}_j,\mathbf{sh}_j\}_{j=1}^{H\cdot W\cdot N}
    \label{eq:bandlimited-gs}
\end{equation}

\subsubsection{Opacity Balancing.}

The 3D-BLPF prevents Gaussian primitives from shrinking smaller than certain minimum scales. This tends to expand the pixel-aligned Gaussian primitives, increasing the overlap between them, which in turn decreases the visibilities of the Gaussian primitives that were positioned behind the expanded Gaussians and were not originally occluded. This degrades the NVS performance, and causes unnatural grid-like artifacts when rendering rate is increased (zoomed-in). To compensate for this effect, we adopt an opacity balancing (OB) technique \cite{he2026surfsplat} that is applied for our band-limited Gaussians to ensure that all pixel-aligned Gaussian primitives are seamlessly integrated into the rendering process. To do so, after 3D-BLPF, the opacities of all 3DGS primitives are clamped using an upper bound threshold as:
\begin{equation}
  \begin{split}
  \alpha_j^\text{clip} = 
  \begin{dcases*} 
  \alpha_j & if  $\alpha_j < \tau_\text{opa}$, \\ 
  \tau_\text{opa} & if $\alpha_j \geq \tau_\text{opa}$
  \end{dcases*} 
  \end{split}
  \label{eq:opacity-thresholding}
\end{equation}
where $\tau_\text{opa}<1$ is an opacity thresholding hyperparameter. This prevents any Gaussian primitive from becoming fully opaque, ensuring that all Gaussians become sufficiently visible and contribute substantially to the rendering regardless of depth order.
Then, we normalize the rendered colors to compensate for the transparency effects that this opacity clamping can cause. Specifically, we also render the alpha values $\alpha(\mathbf{x})$ alongside the intermediate colors $\tilde{C}(\mathbf{x})$ during rasterization according to:
\begin{equation}
  \begin{split}
  \alpha(\mathbf{x})=\sum_{i\in\mathcal{N}}\alpha^\text{clip}_i\mathcal{G}_i^\text{2D}(\mathbf{x})\prod_{j=1}^{i-1}(1-\alpha^\text{clip}_j\mathcal{G}_j^\text{2D}(\mathbf{x})),
  \end{split}
  \label{eq:ob_alpha_rasterization}
\end{equation}
\begin{equation}
  \begin{split}
  \tilde{C}(\mathbf{x})=\sum_{i\in\mathcal{N}}\mathbf{c}_i\alpha^\text{clip}_i\mathcal{G}_i^\text{2D}(\mathbf{x})\prod_{j=1}^{i-1}(1-\alpha^\text{clip}_j\mathcal{G}_j^\text{2D}(\mathbf{x})).
  \end{split}
  \label{eq:ob_color_rasterization}
\end{equation}
We then use the rendered final alpha $\alpha(\mathbf{x})<1$ to normalize the partially transparent colors $\tilde{C}(\mathbf{x})$, as follows:
\begin{equation}
  \begin{split}
  C_\text{OB}(\mathbf{x})=
  \begin{dcases*}
  \tilde{C}(\mathbf{x}) & if $\alpha(\mathbf{x})<\tau_\alpha,$ \\
  \frac{\tilde{C}(\mathbf{x})}{\alpha(\mathbf{x})} & if $\alpha(\mathbf{x})\geq\tau_\alpha$
  \end{dcases*}
  \end{split}
  \label{eq:ob_color_normalization}
\end{equation}
where $\tau_\alpha$ is a threshold hyperparameter that is used for numerical stability. $\alpha(\mathbf{x})<\tau_\alpha$ refers to target view regions that have almost no overlap with any of the context views, and should be left almost transparent. On the other hand, $\alpha(\mathbf{x})\geq\tau_\alpha$ refers to the regions that overlap with at least one context view - as visible geometry is present, these regions should be assumed to be opaque.

\subsection{Optimization}
We train our full model using a weighted sum of mean squared error (MSE) and LPIPS \cite{zhang2018unreasonable} losses between RGB colors of rendered and ground truth target views on a single target resolution:
\begin{equation}
    \mathcal{L}=\sum_{m=1}^{M}\big(\mathcal{L}_\text{MSE}(I_\text{render}^m,I_\text{GT}^m)+\lambda\cdot\mathcal{L}_\text{LPIPS}(I_\text{render}^m,I_\text{GT}^m)\big),
    \label{eq:loss}
\end{equation}
where $M$ is the number of novel views to render for each scene, and the weight for the LPIPS loss term is set to $\lambda=0.05$.

\section{Experimental Results}
\label{sec:results}

\subsection{Experimental Setup}
\label{sec:implementation-details}
\noindent \textbf{Implementation Details.}
We implement our method in PyTorch, based on the DepthSplat \cite{xu2025depthsplat} architecture. We train our model on the RealEstate10K (RE10K) \cite{zhou2018stereo} dataset with 2 context views at $256\times256$ resolution using a total batch size of 8 for 600K steps.
We use the variant that uses the ViT-S monocular backbone for all experiments. Training takes approximately 3 days on 2 NVIDIA RTX 4090 GPUs.
We replace the vanilla 3DGS screen-space dilation filter \cite{kerbl20233d} with the 2D Mip filter \cite{yu2024mip}. Following Mip-splatting \cite{yu2024mip}, we set $s=0.1$ for the 2D Mip filter and $\sigma_s=0.2$ for the 3D Post-filter. Following SurfSplat \cite{he2026surfsplat}, we set $\tau_\text{opa}=0.6$ and set $\tau_\alpha=0.1$ during training and $\tau_\alpha=0.001$ during testing.\\

\noindent \textbf{Datasets.}
We evaluate AA-Splat and existing methods by training on RE10K \cite{zhou2018stereo} and evaluating in-distribution RE10K, as well as zero-shot generalization on DL3DV \cite{ling2024dl3dv} and ACID \cite{liu2021infinite} datasets.
For each scene, each FF-3DGS model under comparison is run once on 2 context views at a single resolution to obtain 3D Gaussian representations. Target views are then rendered and evaluated at multiple resolutions, where rendering at higher resolutions simulates zoom-in effects and rendering at lower resolutions simulates zoom-out effects.\\

\noindent \textbf{Metrics.}
Although the DL3DV \cite{ling2024dl3dv} dataset provides high-resolution test frames, enabling us to simply report PSNR, SSIM, and LPIPS of multi-scale rendered views, RE10K and ACID test sets do not provide higher-resolution test frames. As a proxy, we instead report the High Resolution Rendering Consistency (HRRC) metric \cite{he2026surfsplat} to evaluate upsampling quality in those datasets. This involves bicubic upsampling low-resolution ground truth target view images and using them as reference to compute PSNR, SSIM, or LPIPS \cite{zhang2018unreasonable} metrics:
\begin{equation}
  \begin{split}
  \text{HRRC}_\text{metric}=\text{metric}(\hat{I}^\text{HR}, \hat{I}^\text{GT}), \text{metric}\in\{\text{PSNR}, \text{SSIM}, \text{LPIPS}\},
  \end{split}
  \label{eq:hrrc}
\end{equation}
where $\hat{I}^\text{HR}$ denotes the high-resolution novel view and $\hat{I}^\text{GT}$ refers to the upsampled ground truth target images.

\begin{figure*}[t]
  \centering
  \includegraphics[width=1.0\linewidth]{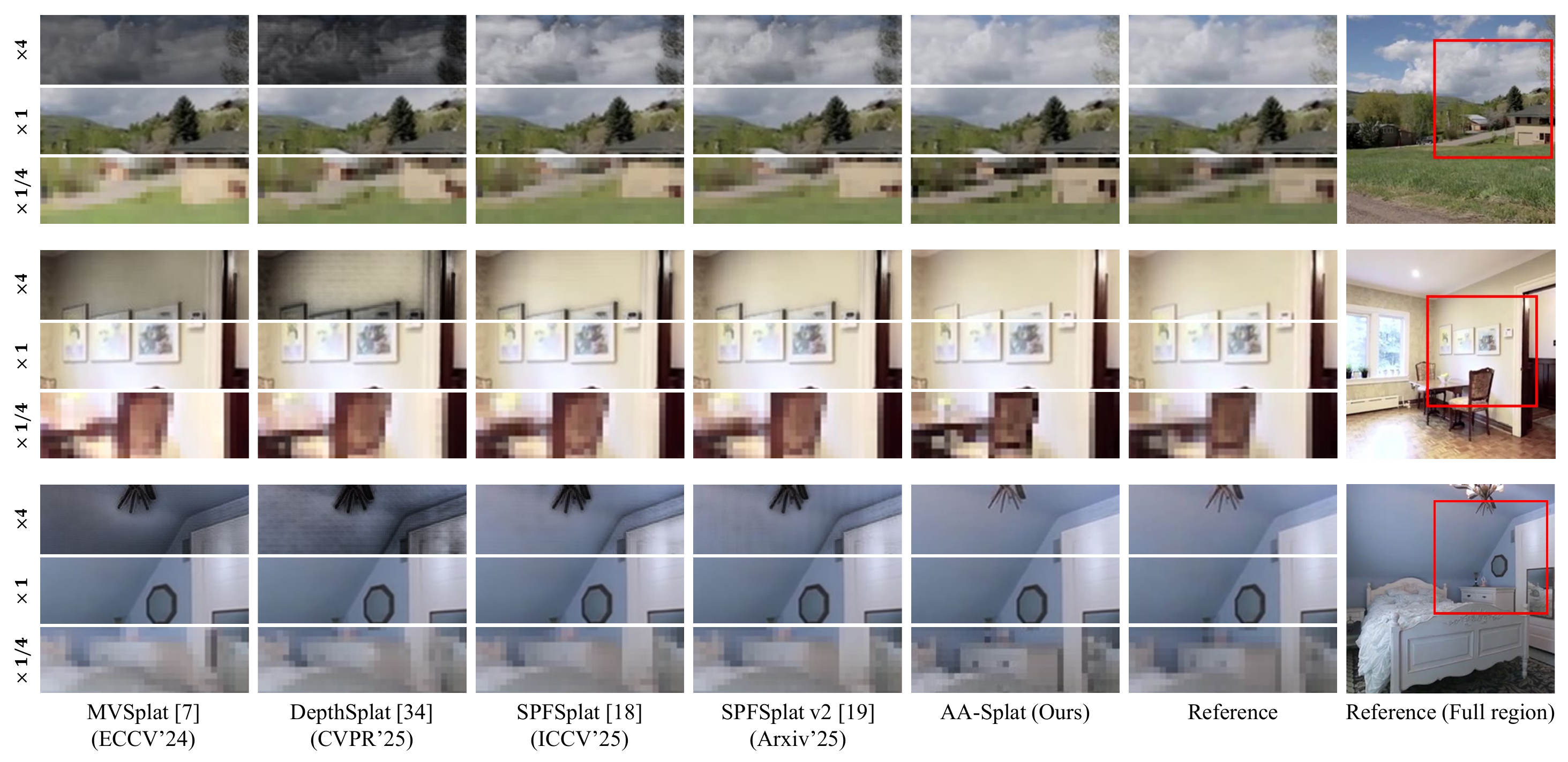}
   \caption{\textbf{Multi-scale visual comparison on RE10K \cite{zhou2018stereo}.} $\times4$ indicates rendering target views at higher resolution to simulate zoom-in effects, while $\times1/4$ indicates rendering target views at lower resolution to simulate zoom-out effects. Additional visualizations will be available in supplementary materials.
   }
   \label{fig:re10k-qualitative}
\end{figure*}

\begin{figure*}[t]
  \centering
  \includegraphics[width=1.0\linewidth]{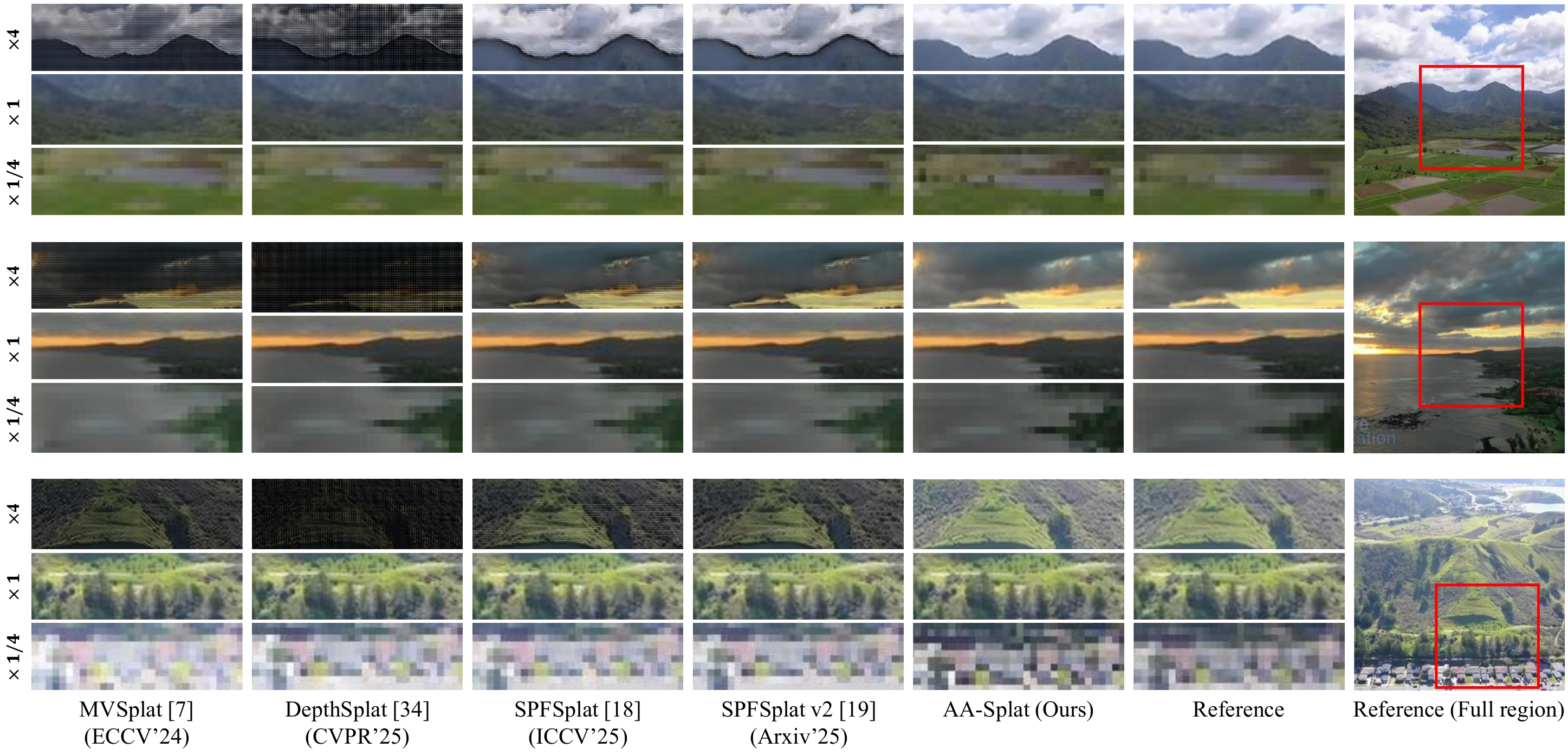}
   \caption{\textbf{Multi-scale visual comparison of cross-dataset generalization on ACID \cite{liu2021infinite}.} Models trained on RE10K are directly used to test on scenes from ACID. Additional visualizations will be available in supplementary materials.}
   \label{fig:acid-qualitative}
\end{figure*}

\begin{table*}
  \caption{\textbf{Quantitative comparison of single-scale reconstruction and multi-scale rendering on the RE10K \cite{zhou2018stereo} dataset under various target resolutions.} We evaluate all models by performing a single feed-forward inference under $256\times256$ input resolution and rendering the resulting 3D Gaussians at $1/4\times$, $1/2\times$, $1\times$, $2\times$, and $4\times$ resolutions. Due to the absence of high-resolution ground truths in the RE10K dataset, we instead report the HRRC metrics which are computed against bicubic upsampled ground truth frames, for $2\times$ and $4\times$ resolutions.
  \textcolor{red}{\textbf{Red}} and \textcolor{blue}{\underline{Blue}} indicate the best and second-best performances, respectively.}
  \label{tab:re10k-256x256-id}
  \centering
  \resizebox{\textwidth}{!}{
      \begin{tabular}{@{}l|cccccc|cccccc|cccccc@{}}
        \toprule
        {Method}  & \multicolumn{6}{c|}{PSNR (dB)$\uparrow$} & \multicolumn{6}{c|}{SSIM$\uparrow$} & \multicolumn{6}{c}{LPIPS$\downarrow$}\\
        & $1/4\times$ Res. & $1/2\times$ Res. & $1\times$ Res. & $2\times$ Res. & $4\times$ Res. & Avg. & $1/4\times$ Res. & $1/2\times$ Res. & $1\times$ Res. & $2\times$ Res. & $4\times$ Res. & Avg. & $1/4\times$ Res. & $1/2\times$ Res. & $1\times$ Res. & $2\times$ Res. & $4\times$ Res. & Avg. \\
        \midrule
        MVSplat \cite{chen2024mvsplat} & 20.252 & 22.428 & 26.387 & 20.294 & 17.882 & 21.449 & 0.735 & 0.847 & 0.869 & 0.808 & 0.753 & 0.803 & 0.152 & 0.129 & 0.128 &  0.279&  0.421 & 0.220\\
        DepthSplat \cite{xu2025depthsplat} &  21.300& 24.449& 26.791 &  20.595&  17.399&  22.107& 0.740& 0.864& 0.878 &  0.779&  0.671&  0.787& 0.139& 0.102& 0.122 &  0.319&  0.449& 0.226\\
        NoPoSplat \cite{ye2024no} & \textcolor{blue}{\underline{21.367}}& \textcolor{blue}{\underline{24.786}}& 26.755 & 21.808 & 18.809 & 22.705 & 0.739 & 0.871 & 0.878 & 0.821 & 0.755 & 0.813 & 0.126 & 0.095 & 0.126 & 0.266 & 0.362 & 0.195\\
        SPFSplat \cite{huang2025no} & 21.283 & 24.694 & 27.313 &  21.929&  18.922&  22.828 & 0.740 & 0.873 & \textcolor{blue}{\underline{0.888}}&  0.827&  0.765&  0.819 & 0.127 & \textcolor{blue}{\underline{0.093}} & 0.119 &  0.247&  0.353& 0.188\\
        SPFSplatV2 \cite{huang2025spfsplatv2} & 21.329& 24.781& \textcolor{red}{\textbf{27.519}} &  22.305&  19.444 & \textcolor{blue}{\underline{23.076}} & \textcolor{blue}{\underline{0.741}} & \textcolor{blue}{\underline{0.876}} & \textcolor{red}{\textbf{0.892}} &  \textcolor{blue}{\underline{0.837}}&  \textcolor{blue}{\underline{0.785}}&  \textcolor{blue}{\underline{0.826}} & \textcolor{blue}{\underline{0.125}} & 0.091& 0.116 & \textcolor{blue}{\underline{0.237}}&  \textcolor{blue}{\underline{0.340}}& \textcolor{blue}{\underline{0.182}}\\
        SPFSplatV2-L \cite{huang2025spfsplatv2} & 20.514 & 24.215 & \textcolor{blue}{\underline{27.426}} & 20.013 & 16.324 & 21.698 &  0.703 &  0.862 & \textcolor{red}{\textbf{0.892}} &  0.782&  0.633&  0.774 &  0.138 & 0.093& \textcolor{red}{\textbf{0.106}} &  0.337&  0.488& 0.233\\
        Ours & \textcolor{red}{\textbf{27.919}}& \textcolor{red}{\textbf{28.110}}& 27.279& \textcolor{red}{\textbf{27.153}}& \textcolor{red}{\textbf{27.087}}& \textcolor{red}{\textbf{27.510}}& \textcolor{red}{\textbf{0.930}}& \textcolor{red}{\textbf{0.915}}& 0.887& \textcolor{red}{\textbf{0.887}}& \textcolor{red}{\textbf{0.904}}& \textcolor{red}{\textbf{0.904}}& \textcolor{red}{\textbf{0.063}}& \textcolor{red}{\textbf{0.079}}& \textcolor{blue}{\underline{0.114}}& \textcolor{red}{\textbf{0.169}}& \textcolor{red}{\textbf{0.200}}& \textcolor{red}{\textbf{0.125}}\\
        \bottomrule
      \end{tabular}
  }
\end{table*}

\begin{table*}[t]
  \caption{\textbf{Zero-shot generalization from RE10K \cite{zhou2018stereo} to DL3DV \cite{ling2024dl3dv} dataset under various target resolutions.} We evaluate all models by performing a single feed-forward inference with 2 context views of $256\times448$ input resolution and rendering the resulting 3D Gaussians at $1/4\times$, $1/2\times$, $1\times$, $2\times$, and $4\times$ resolutions.
  }
  \label{tab:dl3dv-256x448-zs}
  \centering
  \resizebox{\textwidth}{!}{
      \begin{tabular}{@{}l|cccccc|cccccc|cccccc@{}}
        \toprule
        {Method}  & \multicolumn{6}{c|}{PSNR (dB)$\uparrow$} & \multicolumn{6}{c|}{SSIM$\uparrow$} & \multicolumn{6}{c}{LPIPS$\downarrow$}\\
        & $1/4\times$ Res. & $1/2\times$ Res. & $1\times$ Res. & $2\times$ Res. & $4\times$ Res. & Avg. & $1/4\times$ Res. & $1/2\times$ Res. & $1\times$ Res. & $2\times$ Res. & $4\times$ Res. & Avg. & $1/4\times$ Res. & $1/2\times$ Res. & $1\times$ Res. & $2\times$ Res. & $4\times$ Res. & Avg. \\
        \midrule
        DepthSplat \cite{xu2025depthsplat} & 19.687& 22.660& 27.109&  17.294&  13.357&  20.022&  0.679&  0.848&  0.885&  0.647&  0.419&  0.695&  0.173&  0.108&  0.104&  0.399&  0.566& 0.270\\
        Ours & \textbf{28.183}& \textbf{29.012}& \textbf{27.882}&  \textbf{25.701}&  \textbf{24.351}&  \textbf{27.026}&  \textbf{0.934}&  \textbf{0.927}&  \textbf{0.892}&  \textbf{0.821}&  \textbf{0.762}&  \textbf{0.867}&  \textbf{0.061}&  \textbf{0.067}&  \textbf{0.096}&  \textbf{0.233}&  \textbf{0.365}& \textbf{0.164}\\
        \bottomrule
      \end{tabular}
  }
\end{table*}

\begin{table*}[t]
  \caption{\textbf{Zero-shot generalization from RE10K \cite{zhou2018stereo} to ACID \cite{liu2021infinite} dataset under various target resolutions.}
  We evaluate all models by performing a single feed-forward inference under $256\times256$ input resolution and rendering the resulting 3D Gaussians at $1/4\times$, $1/2\times$, $1\times$, $2\times$, and $4\times$ resolutions. Due to the absence of high-resolution ground truths in the ACID dataset, we instead report the HRRC metrics, which are computed against bicubic upsampled ground truth frames, for $2\times$ and $4\times$ resolutions.
  \textcolor{red}{\textbf{Red}} and \textcolor{blue}{\underline{Blue}} indicate the best and second-best performances, respectively.
  }
  \label{tab:acid-256x448-zs}
  \centering
  \resizebox{\textwidth}{!}{
      \begin{tabular}{@{}l|cccccc|cccccc|cccccc@{}}
        \toprule
        {Method}  & \multicolumn{6}{c|}{PSNR (dB)$\uparrow$} & \multicolumn{6}{c|}{SSIM$\uparrow$} & \multicolumn{6}{c}{LPIPS$\downarrow$}\\
        & $1/4\times$ Res. & $1/2\times$ Res. & $1\times$ Res. & $2\times$ Res. & $4\times$ Res. & Avg. & $1/4\times$ Res. & $1/2\times$ Res. & $1\times$ Res. & $2\times$ Res. & $4\times$ Res. & Avg. & $1/4\times$ Res. & $1/2\times$ Res. & $1\times$ Res. & $2\times$ Res. & $4\times$ Res. & Avg. \\
        \midrule
        MVSplat \cite{chen2024mvsplat} & 20.899 & 22.232 & 28.160 & 19.226 & 16.422 & 21.388 & \textcolor{blue}{\underline{0.730}} & 0.829 & 0.841 & 0.745 & 0.648 & 0.759 & 0.154 & 0.141 & 0.147 & 0.349 & 0.507 & 0.259\\
        DepthSplat \cite{xu2025depthsplat} & 22.264& 24.512& 28.026 & 18.019 & 14.078 & 21.380& 0.701& 0.823& 0.841 & 0.622 & 0.421 & 0.682& 0.145& 0.126& 0.147 & 0.486 & 0.624 & 0.306\\
        NoPoSplat \cite{ye2024no} & \textcolor{blue}{\underline{22.843}} & 25.184 & 27.452 & 21.188 & 18.060 & 22.945 & 0.723 & 0.832 & 0.824 & 0.755 & 0.674 & 0.761 & 0.130 & 0.120 & 0.161 & 0.345 & 0.452 & 0.242\\
        SPFSplat \cite{huang2025no} & 22.532 & 24.814 & 28.077 & 21.217 & 18.155 & 22.959 & 0.725 & 0.842 & 0.841 & 0.765 & 0.688 & 0.772 & 0.129 & 0.115 & \textcolor{red}{\textbf{0.131}} & 0.311 & 0.435 & 0.228\\
        SPFSplatV2 \cite{huang2025spfsplatv2} & 22.734 & 25.159 & 28.322 & \textcolor{blue}{\underline{21.714}}& \textcolor{blue}{\underline{18.822}}& \textcolor{blue}{\underline{23.350}} & 0.726 & \textcolor{blue}{\underline{0.845}} & 0.846 & \textcolor{blue}{\underline{0.781}}& \textcolor{blue}{\underline{0.721}} & \textcolor{blue}{\underline{0.784}} & \textcolor{blue}{\underline{0.125}} & 0.111& 0.145 & \textcolor{blue}{\underline{0.296}}& \textcolor{blue}{\underline{0.418}} & \textcolor{blue}{\underline{0.219}} \\
        SPFSplatV2-L \cite{huang2025spfsplatv2} & 22.400 & \textcolor{blue}{\underline{25.299}} & \textcolor{blue}{\underline{28.398}} & 19.142 & 15.360 & 22.120 & 0.698 & 0.837 & \textcolor{red}{\textbf{0.851}} & 0.697 & 0.499 & 0.716 & 0.129 & \textcolor{blue}{\underline{0.107}}& \textcolor{red}{\textbf{0.131}}& 0.451 & 0.597 & 0.283 \\
        Ours & \textcolor{red}{\textbf{29.672}} & \textcolor{red}{\textbf{29.689}} & \textcolor{red}{\textbf{28.483}} & \textcolor{red}{\textbf{28.287}} & \textcolor{red}{\textbf{28.183}} & \textcolor{red}{\textbf{28.863}} & \textcolor{red}{\textbf{0.904}} & \textcolor{red}{\textbf{0.885}} & \textcolor{blue}{\underline{0.847}} & \textcolor{red}{\textbf{0.851}} & \textcolor{red}{\textbf{0.876}} & \textcolor{red}{\textbf{0.873}} & \textcolor{red}{\textbf{0.081}} & \textcolor{red}{\textbf{0.103}}& \textcolor{blue}{\underline{0.141}}& \textcolor{red}{\textbf{0.201}} & \textcolor{red}{\textbf{0.243}} & \textcolor{red}{\textbf{0.154}}\\
        \bottomrule
      \end{tabular}
  }
\end{table*}

\subsection{Comparisons on RE10K}
Table \ref{tab:re10k-256x256-id} and Fig. \ref{fig:re10k-qualitative} present quantitative and qualitative comparisons of FF-3DGS models under varying target resolutions. We compare our AA-Splat with the baseline DepthSplat \cite{xu2025depthsplat}, as well as representative FF-3DGS models such as MVSplat \cite{chen2024mvsplat}, NoPoSplat \cite{ye2024no}, and SPFSplat \cite{huang2025no,huang2025spfsplatv2}.
Most existing state-of-the-art FF-3DGS models are entirely incapable of handling drastic changes to sampling rates, showing severe degradations when rendering at increased or decreased resolutions. Under zoom-in, erosion artifacts open up seams between pixel-aligned primitives, manifesting as dark regions. Under zoom-out, dilation artifacts cause unnaturally thick structures to be rendered.
In contrast, our AA-Splat maintains high rendering fidelity over a wide range of target resolutions, with only marginal degradation even under aggressive zoom-in/out factors.
These results highlight the robustness and effectiveness of our OBBL design.

\begin{figure*}
  \centering
  \includegraphics[width=1.0\linewidth]{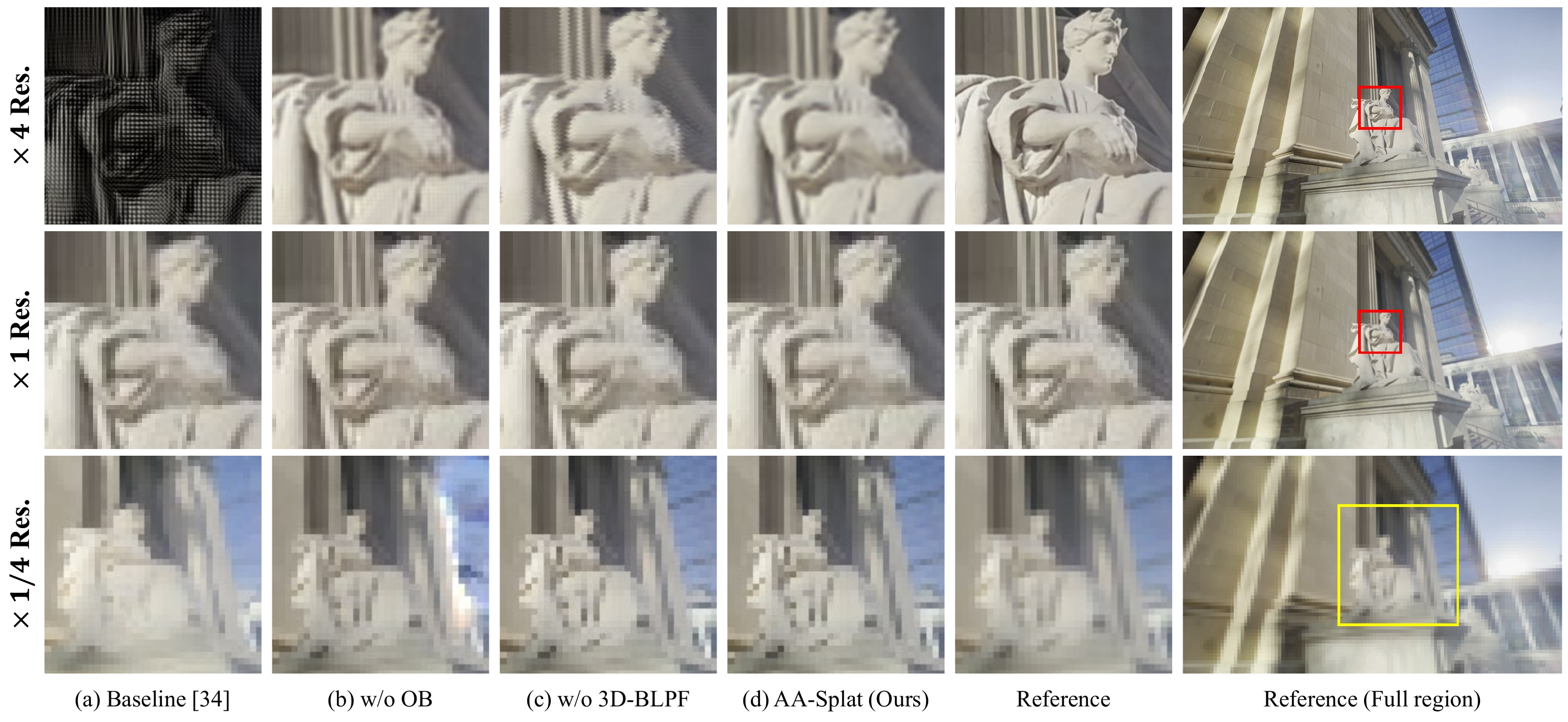}
   \caption{\textbf{Ablation results on the DL3DV \cite{ling2024dl3dv} dataset.} All methods are trained and inferred at full resolution and rendered at $4\times$ resolution to mimic zoom-in.
   Removing either OB or 3D-BLPF causes unnatural grid-like artifacts or sharp jagged edges.
   Our method (AA-Splat) effectively suppresses these rendering artifacts.}
   \label{fig:abl-dl3dv-qualitative}
\end{figure*}

\begin{table*}
  \caption{\textbf{Ablation results on the DL3DV \cite{ling2024dl3dv} dataset.} We evaluate all variants by performing a single feed-forward inference under $256\times448$ input resolution and rendering the resulting 3D Gaussians at $1/4\times$, $1/2\times$, $1\times$, $2\times$, and $4\times$ resolutions.
  \textcolor{red}{\textbf{Red}} and \textcolor{blue}{\underline{Blue}} indicate the best and second-best performances, respectively.}
  \label{tab:abl-dl3dv-256x448-zs}
  \centering
  \resizebox{\textwidth}{!}{
      \begin{tabular}{@{}l|cccccc|cccccc|cccccc@{}}
        \toprule
        Variants & \multicolumn{6}{c|}{PSNR (dB)$\uparrow$} & \multicolumn{6}{c|}{SSIM$\uparrow$} & \multicolumn{6}{c}{LPIPS$\downarrow$}\\
         & $1/4$ Res. & $1/2$ Res. & $1\times$ Res. & $2\times$ Res. & $4\times$ Res. & Avg. & $1/4$ Res. & $1/2$ Res. & $1\times$ Res. & $2\times$ Res. & $4\times$ Res. & Avg. & $1/4$ Res. & $1/2$ Res. & $1\times$ Res. & $2\times$ Res. & $4\times$ Res. & Avg. \\
        \midrule
        (a) Baseline \cite{xu2025depthsplat} &  19.687&  22.660
&  27.109&  17.294&  13.357&  20.022
&  0.679&  0.848
&  \textcolor{blue}{\underline{0.885}}&  0.647&  0.419&  0.695
&  0.173&  0.108
&  \textcolor{blue}{\underline{0.104}}&  0.399&  0.566& 0.270
\\ 
        \midrule
        (b) w/o OB &  \textcolor{blue}{\underline{28.140}}&  27.936
&  26.623&  24.868&  23.715&  26.257
&  \textcolor{red}{\textbf{0.934}}&  0.915
&  0.875&  \textcolor{blue}{\underline{0.806}}&  \textcolor{blue}{\underline{0.752}}&  0.856
&  \textcolor{red}{\textbf{0.057}}&  0.076
&  0.113&  \textcolor{blue}{\underline{0.248}}&  \textcolor{blue}{\underline{0.375}}& \textcolor{blue}{\underline{0.174}}\\ 
        (c) w/o 3D-BLPF &  28.013&  \textcolor{blue}{\underline{28.639}}&  \textcolor{blue}{\underline{27.322}}&  \textcolor{blue}{\underline{25.186}}&  \textcolor{blue}{\underline{23.771}}&  \textcolor{blue}{\underline{26.586}}&  \textcolor{blue}{\underline{0.932}}&  \textcolor{blue}{\underline{0.924}}&  \textcolor{blue}{\underline{0.885}}&  \textcolor{blue}{\underline{0.806}}&  0.738&  \textcolor{blue}{\underline{0.857}}&  0.062&  \textcolor{blue}{\underline{0.069}}&  0.112&  0.273&  0.423& 0.188
\\
        \midrule
        (d) Ours &  \textcolor{red}{\textbf{28.183}}&  \textcolor{red}{\textbf{29.012}}&  \textcolor{red}{\textbf{27.882}}&  \textcolor{red}{\textbf{25.701}}&  \textcolor{red}{\textbf{24.351}}&  \textcolor{red}{\textbf{27.026}}&  \textcolor{red}{\textbf{0.934}}&  \textcolor{red}{\textbf{0.927}}&  \textcolor{red}{\textbf{0.892}}&  \textcolor{red}{\textbf{0.821}}&  \textcolor{red}{\textbf{0.762}}&  \textcolor{red}{\textbf{0.867}}&  \textcolor{blue}{\underline{0.061}}&  \textcolor{red}{\textbf{0.067}}&  \textcolor{red}{\textbf{0.096}}&  \textcolor{red}{\textbf{0.233}}&  \textcolor{red}{\textbf{0.365}}& \textcolor{red}{\textbf{0.164}}\\
        \bottomrule
      \end{tabular}
  }
\end{table*}

\subsection{Cross-Dataset Generalization}
\noindent Tables \ref{tab:dl3dv-256x448-zs} and \ref{tab:acid-256x448-zs} and Fig. \ref{fig:acid-qualitative} show the zero-shot cross-dataset generalization performances of AA-Splat to the unseen DL3DV \cite{ling2024dl3dv} and ACID \cite{liu2021infinite} datasets, respectively, compared against recent SOTA FF-3DGS models. Models trained on RE10K \cite{zhou2018stereo} with 2 context views of $256\times256$ are evaluated directly on the DL3DV \cite{ling2024dl3dv} dataset at 2 context views of $256\times448$ resolution, and on the ACID \cite{liu2021infinite} dataset at 2 context views of $256\times256$ resolution, without any finetuning. Most existing models exhibit severe degradations at out-of-distribution sampling rates on the ACID \cite{liu2021infinite} dataset, exhibiting similarly severe erosion and dilation artifacts. In the DL3DV \cite{ling2024dl3dv} dataset, the baseline DepthSplat shows even greater performance drops under zoom-ins than on RE10K \cite{zhou2018stereo} or ACID \cite{liu2021infinite}. In contrast, our AA-Splat maintains robust performance across all resolutions on both datasets. Averaged across all resolutions, AA-Splat achieves a \textit{substantial} improvement of 7.004dB in the DL3DV \cite{ling2024dl3dv} dataset and 7.483dB in the ACID \cite{liu2021infinite} dataset, compared to the baseline DepthSplat \cite{xu2025depthsplat}.
Note that AA-Splat also shows significant improvements with +0.773dB in DL3DV \cite{ling2024dl3dv} and +0.457dB in ACID \cite{liu2021infinite} over the baseline DepthSplat \cite{xu2025depthsplat} in zero-shot NVS at the native ($1\times$) target resolution as well.

\subsection{Ablation and Analysis}

We assess the contribution of each component via ablation studies, as shown in Table \ref{tab:abl-dl3dv-256x448-zs} and Fig. \ref{fig:abl-dl3dv-qualitative}. We train all variants on RE10K \cite{zhou2018stereo} and test on the DL3DV \cite{ling2024dl3dv} dataset, due to the presence of high-resolution ground truths in DL3DV \cite{ling2024dl3dv}.

\noindent \textbf{Opacity Balancing (OB).} Comparing (b) and (d) in Table \ref{tab:abl-dl3dv-256x448-zs} shows that removing OB and leaving only the 3D-BLPF in (b) degrades NVS performance significantly across all resolutions. Notably, comparing (a) and (b) in Table \ref{tab:abl-dl3dv-256x448-zs} shows that this setup also degrades the native resolution ($1\times$) NVS performance compared to the baseline DepthSplat \cite{xu2025depthsplat}. Comparing (b) and (d) in Fig. \ref{fig:abl-dl3dv-qualitative} shows that the 3D-BLPF on its own introduces unnatural grid-like artifacts upon zoom-in. These artifacts were not present in previous per-scene anti-aliased 3DGS methods \cite{yu2024mip, younes2025anti}. These results highlight the decreased visibility problems that arise from the increase in overlap between expanded Gaussian primitives, and the necessity of OB compensation.\\
\noindent \textbf{3D Band-Limiting Post-Filtering (3D-BLPF).} Comparing (c) and (d) in Table \ref{tab:abl-dl3dv-256x448-zs} and Fig. \ref{fig:abl-dl3dv-qualitative}, it can be noted that the removal of the 3D-BLPF in (c) causes larger drops in NVS performance under higher resolutions ($2\times$, $4\times$) than it does at the native ($1\times$) or lower resolutions ($1/2\times$, $1/4\times$). Comparing (c) and (d) in Fig. \ref{fig:abl-dl3dv-qualitative} reveals that the removal of 3D-BLPF introduces high-frequency artifacts in the form of jagged edges. These results demonstrate that the role of 3D-BLPF is more prominent when rendering at increased sampling factors, and highlight the need for explicit band-limiting of 3D Gaussian primitives for suppressing high-frequency artifacts during zoom-in.

\section{Conclusion}
We present AA-Splat, a feedforward 3DGS model which \textbf{firstly} incorporates anti-aliasing designs into its architecture to enable alias-free rendering at arbitrary sampling rates deviating heavily from those present in the context views.
In order to effectively regenerate the degenerate Gaussians, we propose to apply a 3D band-limiting post-filter, which  effectively bandlimits the 3D Gaussian primitives according to the Nyquist sampling rate, thus eliminating high-frequency artifacts from degenerate Gaussians, while the opacity balancing (OB) technique effectively compensates for the resulting increase in overlap between pixel-aligned primitives. Both techniques complement each other effectively, resulting not only in an effective anti-aliasing solution, but also a significant improvement in cross-domain generalization capability.
Our experimental results demonstrate across standard feedforward novel view synthesis benchmarks that AA-Splat consistently and vastly outperforms the original DepthSplat (a baseline) \cite{xu2025depthsplat} with \textbf{significant margins of average 5.4 $\sim$ 7.5 dB PSNR} when reconstructing and rendering at varied out-of-distribution sampling rates (e.g., zooming in/out), while also improving NVS performance in the native resolution.

\section{Acknowledgements}
This work was supported by IITP grant funded by the Korea government (MSIT) (No.RS2022-00144444, Deep Learning Based Visual Representational Learning and Rendering of Static and Dynamic Scenes).

\clearpage









\title{AA-Splat: Supplementary Material}



\author{}
\institute{}

\maketitle


\section{Screen-space Dilation Filters}
\textbf{Vanilla 3DGS Dilation Filter.} The original 3DGS \cite{kerbl20233d} handled degenerate Gaussians cases, where projected 2D Gaussians were smaller than a single pixel, by employing a screen-space dilation filter to forcibly enlarge the projected Gaussians in image space and ensure adequate coverage for rasterization:
\begin{equation}
   \begin{split}
   \mathcal{G}_k^\text{2D}(\mathbf{x}^\text{2D})_\text{dilated} = \exp{\bigg(-\frac{1}{2}(\mathbf{x}^\text{2D}-\bm{\upmu}_k^\text{2D})^T(\mathbf{\Sigma}_k^\text{2D}+s\mathbf{I})^{-1}(\mathbf{x}^\text{2D}-\bm{\upmu}_k^\text{2D})\bigg)},
   \end{split}
   \label{eq:2d_dilation_filter}
\end{equation}
However, this dilation filter caused the optimization process to learn incorrect Gaussian scales, resulting in aliasing artifacts when sampling rate was altered from those visible during training.

\noindent \textbf{2D Mip Filter.} The 2D screen-space Mip filter \cite{yu2024mip} is a screen-space dilation filter that simulates the physical imaging process of integrating photons into a pixel on the camera sensor over the pixel area. In the context of 3DGS rasterization, this can be done by convolving a 2D box filter that has the size of a single pixel with the projected 2D Gaussians. For efficiency, this is further approximated by instead convolving a 2D Gaussian filter with the Gaussian's envelope size (scale) chosen specifically to mimic a single pixel as:
\begin{equation}
   \begin{split}
   \mathcal{G}_k^\text{2D}(\mathbf{x}^\text{2D})_\text{mip} = \sqrt{\frac{|\mathbf{\Sigma}_k^\text{2D}|}{|\mathbf{\Sigma}_k^\text{2D}+s\mathbf{I}|}}e^{\big(-\frac{1}{2}(\mathbf{x}^\text{2D}-\bm{\upmu}_k^\text{2D})^T(\mathbf{\Sigma}_k^\text{2D}+s\mathbf{I})^{-1}(\mathbf{x}^\text{2D}-\bm{\upmu}_k^\text{2D})\big)},
   \end{split}
   \label{eq:2d_mip_filter}
\end{equation}
where $s$ is set to $0.1$ to cover approximately a single pixel in the image space. This operation, unlike the vanilla 3DGS dilation filter, also has the effect of attenuating the opacity of dilated Gaussian primitives by a view-dependent factor $\sqrt{|\mathbf{\Sigma}_k^\text{2D}|/|\mathbf{\Sigma}_k^\text{2D}+s\mathbf{I}|}$.

\noindent \textbf{Dilation Filters in FF-3DGS.} Most existing FF-3DGS models \cite{chen2024mvsplat, xu2025depthsplat, ye2024no, huang2025no, huang2025spfsplatv2} were built on the vanilla 3DGS dilation filter, which was the main reason behind the severe aliasing artifacts. As the regressed 3D Gaussians themselves were highly degenerate (i.e., smaller than a single screen-space pixel), increasing the sampling rate (e.g., zooming in, rendering at higher resolutions) caused erosion artifacts to open up gaps between neighboring pixel-aligned Gaussian primitives. Due to the lack of the opacity attenuation factor and the excessively large dilation scale, decreasing the sampling rate (e.g., zooming out, rendering at lower resolutions) caused dilation artifacts where structures are rendered as being unnaturally thick, as well as brightening artifacts that caused views to be rendered as being unnaturally bright.

\section{Additional Experiments}

\subsection{Additional Comparisons on Zero-Shot DL3DV}
In Table \ref{tab:dl3dv-256x448-zs-supp}, we additionally present the comparison with MVSplat \cite{chen2024mvsplat}. All models were trained on RE10K \cite{zhou2018stereo} and evaluated on test scenes from DL3DV \cite{ling2024dl3dv} without any finetuning. Similarly to DepthSplat, MVSplat exhibits severe performance degradation at all out-of-distribution rendering scales.

\begin{table*}[t]
  \caption{\textbf{Zero-shot generalization from RE10K \cite{zhou2018stereo} to DL3DV \cite{ling2024dl3dv} dataset under various target resolutions.} We evaluate all models by performing a single feed-forward inference with 2 context views of $256\times448$ input resolution and rendering the resulting 3D Gaussians at $1/4\times$, $1/2\times$, $1\times$, $2\times$, and $4\times$ resolutions.
  }
  \label{tab:dl3dv-256x448-zs-supp}
  \centering
  \resizebox{\textwidth}{!}{
      \begin{tabular}{@{}l|cccccc|cccccc|cccccc@{}}
        \toprule
        {Method}  & \multicolumn{6}{c|}{PSNR (dB)$\uparrow$} & \multicolumn{6}{c|}{SSIM$\uparrow$} & \multicolumn{6}{c}{LPIPS$\downarrow$}\\
        & $1/4\times$ Res. & $1/2\times$ Res. & $1\times$ Res. & $2\times$ Res. & $4\times$ Res. & Avg. & $1/4\times$ Res. & $1/2\times$ Res. & $1\times$ Res. & $2\times$ Res. & $4\times$ Res. & Avg. & $1/4\times$ Res. & $1/2\times$ Res. & $1\times$ Res. & $2\times$ Res. & $4\times$ Res. & Avg. \\
        \midrule
        MVSplat \cite{chen2024mvsplat}&  18.690&  19.771&  21.834&  18.173&  16.140&  18.922&  0.635&  0.713&  0.711&  0.630&  0.582&  0.654&  0.201&  0.181&  0.206&  0.349&  0.485& 0.284\\
        DepthSplat \cite{xu2025depthsplat} & 19.687& 22.660& 27.109&  17.294&  13.357&  20.022&  0.679&  0.848&  0.885&  0.647&  0.419&  0.695&  0.173&  0.108&  0.104&  0.399&  0.566& 0.270\\
        Ours & \textbf{28.183}& \textbf{29.012}& \textbf{27.882}&  \textbf{25.701}&  \textbf{24.351}&  \textbf{27.026}&  \textbf{0.934}&  \textbf{0.927}&  \textbf{0.892}&  \textbf{0.821}&  \textbf{0.762}&  \textbf{0.867}&  \textbf{0.061}&  \textbf{0.067}&  \textbf{0.096}&  \textbf{0.233}&  \textbf{0.365}& \textbf{0.164}\\
        \bottomrule
      \end{tabular}
  }
\end{table*}

\subsection{Additional Ablations}
\begin{figure*}
  \centering
  \includegraphics[width=1.0\linewidth]{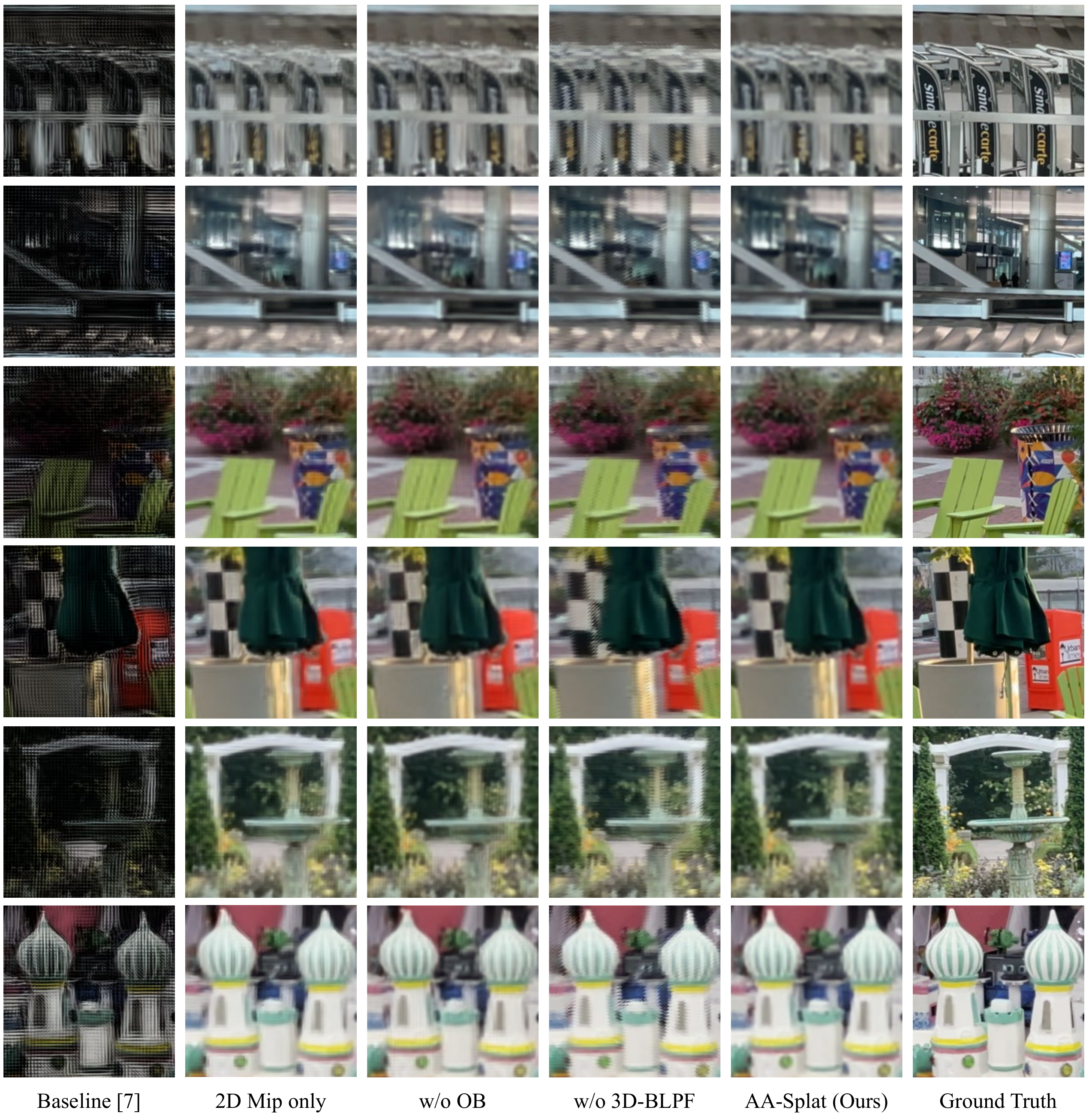}
   \caption{\textbf{Ablation results on the DL3DV \cite{ling2024dl3dv} dataset.} All variants perform feed-forward reconstruction with 2 context views of $256\times448$ resolution and are rendered at $4\times$ resolution to mimic zoom-in effects.
   Removing either OB or 3D-BLPF causes unnatural grid-like artifacts or sharp jagged edges. Using only the 2D Mip filter also causes grid-like artifacts.
   Our method (AA-Splat) effectively suppresses these rendering artifacts, enabling smooth interpolation. Best viewed zoomed-in.}
   \label{fig:abl-dl3dv-qualitative-supp}
\end{figure*}

In Table \ref{tab:abl-dl3dv-256x448-zs-supp} and Fig. \ref{fig:abl-dl3dv-qualitative-supp}, we present additional ablation results on zero-shot DL3DV \cite{ling2024dl3dv}. `2D Mip only' ((d) in Table \ref{tab:abl-dl3dv-256x448-zs-supp}) indicates a retrained variant of DepthSplat where the vanilla 3DGS \cite{kerbl20233d} dilation filter was simply swapped out with the 2D Mip filter \cite{yu2024mip}. `3D-BLPF only' ((e) in Table \ref{tab:abl-dl3dv-256x448-zs-supp}) indicates a variant where only the 3D-BLPF was used, while retaining the vanilla 3DGS dilation filter.

\noindent\textbf{Individual components.}
Comparing (a) and (d) shows that, although the 2D Mip filter greatly mitigates the performance drops that occurred when rendering the baseline model at out-of-distribution sampling rates, it causes a slight drop in the native resolution ($1\times$) NVS performance, and leaves high-frequency artifacts as shown in Fig. \ref{fig:abl-dl3dv-qualitative-supp}. This shows that the 2D Mip filter alone is insufficient to enable alias-free rendering.
Compared with the baseline DepthSplat \cite{xu2025depthsplat}, `3D-BLPF only' shows modest improvement at higher ($2\times, 4\times$) resolutions, but only marginal improvements at lower ($1/2\times, 1/4\times$) resolutions and a slight drop in the $1\times$ resolution. This indicates that the 3D-BLPF primarily targets erosion artifacts during upsampling, but is limited by the increased overlap between pixel-aligned Gaussian primitives.

\noindent\textbf{Component interactions.} As discussed in the main paper, `w/o OB' and `w/o 3D-BLPF' show that 3D-BLPF and OB both complement each other to enable alias-free renderings at increased sampling rates. Additionally, comparing (f) against (g) shows that our complete OBBL performs surprisingly well at most scaling factors even with the flawed vanilla 3DGS dilation filter. At $1/2\times$, $1\times$, $2\times$, and $4\times$ resolutions, it performs comparably to or slightly outperforms our full model using the 2D Mip filter, failing only in the aggressive zoom-out ($1/4\times$ resolution). This shows the strength of our OBBL design, and that it is not dependent on the 2D Mip filter.

\begin{table*}
  \caption{\textbf{Additional ablation results on the DL3DV \cite{ling2024dl3dv} dataset.} We evaluate all variants by performing a single feed-forward inference under $256\times448$ input resolution and rendering the resulting 3D Gaussians at $1/4\times$, $1/2\times$, $1\times$, $2\times$, and $4\times$ resolutions.
  \textcolor{red}{\textbf{Red}} and \textcolor{blue}{\underline{Blue}} indicate the best and second-best performances, respectively.}
  \label{tab:abl-dl3dv-256x448-zs-supp}
  \centering
  \resizebox{\textwidth}{!}{
      \begin{tabular}{@{}l|ccc|cccccc|cccccc|cccccc@{}}
        \toprule
        Variants & 2D- & 3D- & & \multicolumn{6}{c|}{PSNR (dB)$\uparrow$} & \multicolumn{6}{c|}{SSIM$\uparrow$} & \multicolumn{6}{c}{LPIPS$\downarrow$}\\
         & Mip & BLPF & OB & $1/4\times$ & $1/2\times$ & $1\times$ & $2\times$ & $4\times$ & Avg. & $1/4\times$ & $1/2\times$ & $1\times$ & $2\times$ & $4\times$ & Avg. & $1/4\times$ & $1/2\times$ & $1\times$ & $2\times$ & $4\times$ & Avg. \\
        \midrule
        (a) Baseline \cite{xu2025depthsplat} & \xmark & \xmark & \xmark & 19.687&  22.660
&  27.109&  17.294&  13.357&  20.022
&  0.679&  0.848
&  0.885&  0.647&  0.419&  0.695
&  0.173&  0.108
&  0.104&  0.399&  0.566& 0.270
\\
        \midrule
        (b) w/o OB & \cmark & \cmark & \xmark & \textcolor{blue}{\underline{28.140}}&  27.936
&  26.623&  24.868&  23.715&  26.257
&  \textcolor{red}{\textbf{0.934}}&  0.915
&  0.875&  0.806&  0.752&  0.856
&  \textcolor{red}{\textbf{0.057}}&  0.076
&  0.113&  0.248&  0.375& 0.174\\
        (c) w/o 3D-BLPF & \cmark & \xmark & \cmark & 28.013&  28.639&  27.322&  25.186&  23.771&  26.586&  \textcolor{blue}{\underline{0.932}}&  0.924&  0.885&  0.806&  0.738&  0.857&  0.062&  0.069&  0.112&  0.273&  0.423& 0.188
\\
        \midrule
        (d) 2D Mip only & \cmark & \xmark & \xmark & 28.072&  28.066&  27.002&  25.116&  23.886&  26.429&  \textcolor{red}{\textbf{0.934}}&  0.916&  0.879&  0.808&  0.749&  0.857&  \textcolor{blue}{\underline{0.058}}&  0.075&  0.110&  0.241&  0.372& 0.171\\
        (e) 3D-BLPF only & \xmark & \cmark & \xmark & 19.812&  22.698&  26.967&  21.554&  19.162&  22.038&  0.708&  0.861&  0.884&  0.795&  0.717&  0.793&  0.160&  0.102&  0.106&  0.256&  0.400& 0.205
\\
        \midrule
        (f) Ours w/o 2D Mip & \xmark & \cmark & \cmark & 25.703&  \textcolor{red}{\textbf{29.654}}&  \textcolor{blue}{\underline{27.876}}&  \textcolor{red}{\textbf{25.821}}&  \textcolor{red}{\textbf{24.444}}&  \textcolor{blue}{\underline{26.700}}&  0.885&  \textcolor{red}{\textbf{0.931}}&  \textcolor{blue}{\underline{0.890}}&  \textcolor{red}{\textbf{0.828}}&  \textcolor{red}{\textbf{0.770}}&  \textcolor{blue}{\underline{0.861}}&  0.079&  \textcolor{red}{\textbf{0.057}}&  \textcolor{blue}{\underline{0.099}}&  \textcolor{red}{\textbf{0.222}}&  \textcolor{red}{\textbf{0.354}}& \textcolor{red}{\textbf{0.162}}\\
        (g) Ours & \cmark & \cmark & \cmark & \textcolor{red}{\textbf{28.183}}&  \textcolor{blue}{\underline{29.012}}&  \textcolor{red}{\textbf{27.882}}&  \textcolor{blue}{\underline{25.701}}&  \textcolor{blue}{\underline{24.351}}&  \textcolor{red}{\textbf{27.026}}&  \textcolor{red}{\textbf{0.934}}&  \textcolor{blue}{\underline{0.927}}&  \textcolor{red}{\textbf{0.892}}&  \textcolor{blue}{\underline{0.821}}&  \textcolor{blue}{\underline{0.762}}&  \textcolor{red}{\textbf{0.867}}&  0.061&  \textcolor{blue}{\underline{0.067}}&  \textcolor{red}{\textbf{0.096}}&  \textcolor{blue}{\underline{0.233}}&  \textcolor{blue}{\underline{0.365}}& \textcolor{blue}{\underline{0.164}}\\%
        \bottomrule
      \end{tabular}
  }
\end{table*}

\section{Additional Qualitative Results}
In this section, we present further visual comparisons.

\subsection{Additional Visuals on RE10K}
\begin{figure*}
  \centering
  \includegraphics[width=1.0\linewidth]{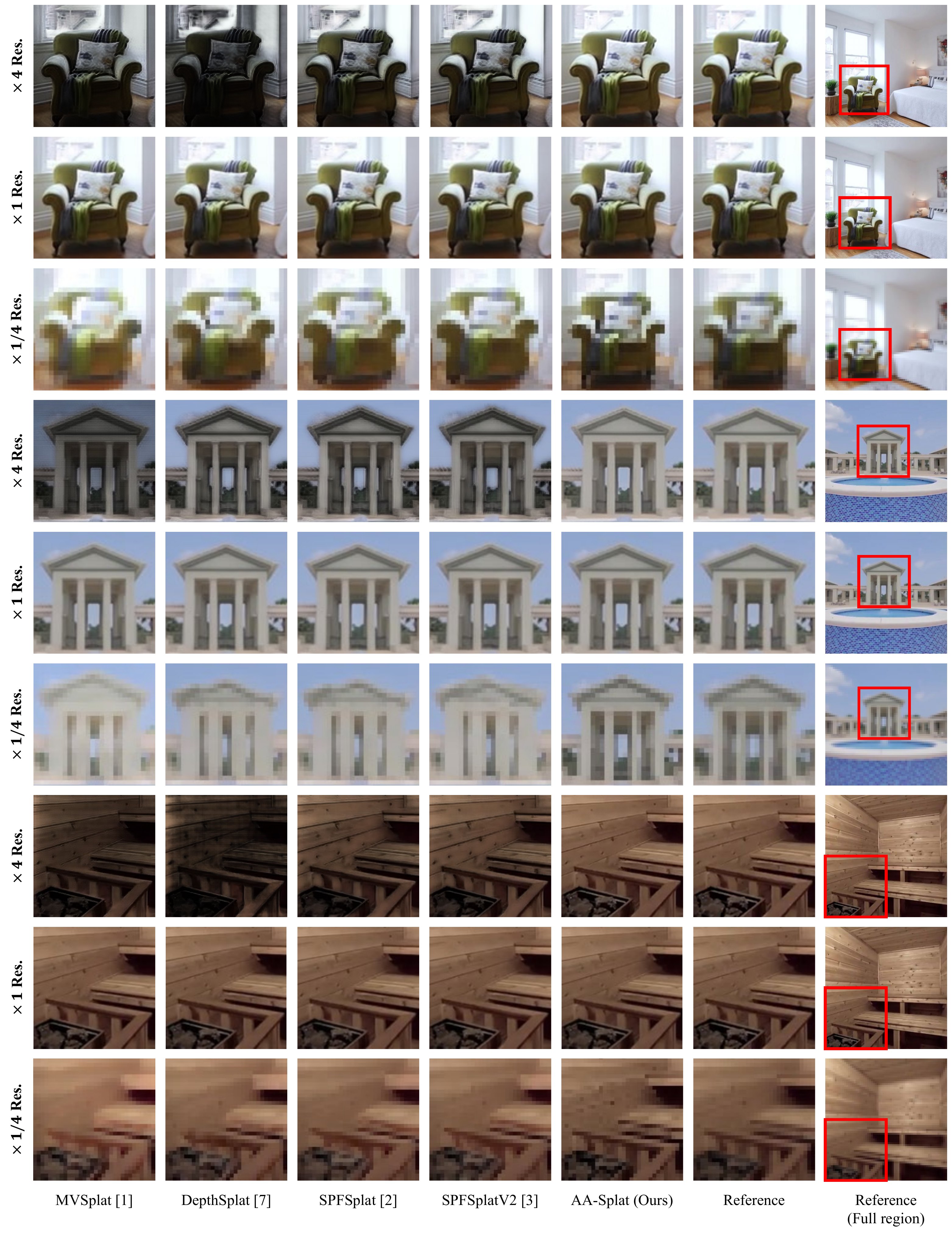}
   \caption{\textbf{Additional multi-scale visual comparisons on RE10K \cite{zhou2018stereo}.} All methods are trained and inferred at $256\times256$ resolution and rendered at $1/4\times \sim 4\times$ resolutions to mimic zoom-in/out effects.
   Best viewed zoomed-in.}
   \label{fig:re10k-qualitative-supp}
\end{figure*}

We present additional visual comparisons with SOTA models on the RE10K \cite{zhou2018stereo} dataset, in Fig. \ref{fig:re10k-qualitative-supp}. Evaluation protocols are identical to the main paper.

\subsection{Additional Visuals on ACID}
\begin{figure*}[t]
  \centering
  \includegraphics[width=0.9\linewidth]{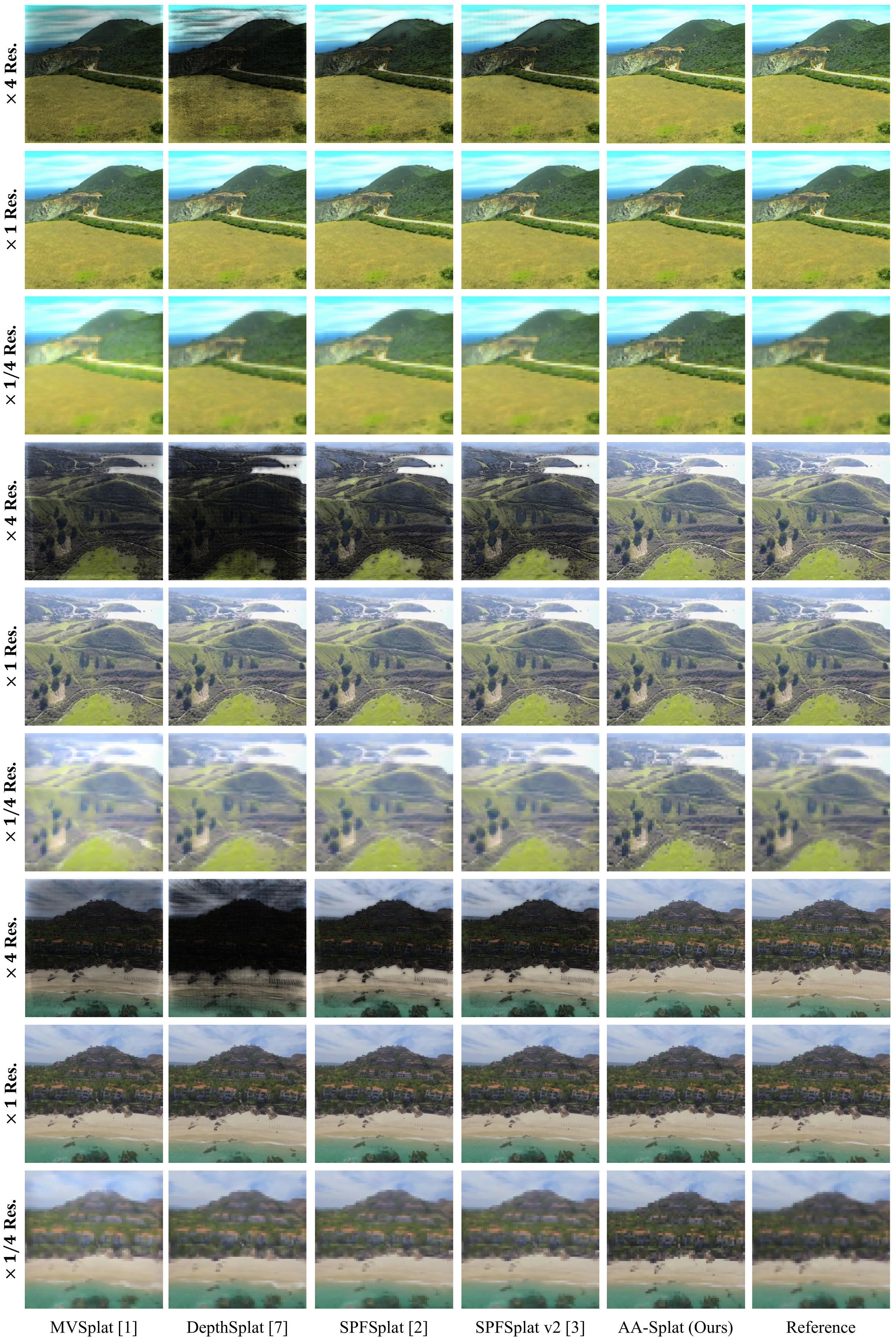}
   \caption{\textbf{Additional multi-scale visual comparisons of cross-dataset generalization to ACID \cite{liu2021infinite}.} Models trained on RE10K are directly tested on scenes from ACID. When rendering novel views at decreased sampling rates on the ACID dataset where thin foreground structures are relatively rare, aliasing artifacts in existing models primarily manifest as brightening artifacts, where scenes are rendered as being unnaturally bright.}
   \label{fig:acid-qualitative-supp}
\end{figure*}

We present additional visual comparisons with SOTA models on the ACID \cite{liu2021infinite} dataset, in Fig. \ref{fig:acid-qualitative-supp}. Evaluation protocols are identical to the main paper, with models trained on RE10K being evaluated on ACID in a zero-shot manner. Note that, due to thin structures being relatively rare in the ACID dataset, dilation artifacts primarily manifest as brightening artifacts, where scenes are rendered as being unnaturally bright.

\clearpage



%
%
\bibliographystyle{splncs04}
\bibliography{main}

\end{document}